\documentclass[10pt]{article}

\usepackage[letterpaper, margin=1in]{geometry}

\usepackage{times}           
\usepackage{mathptmx}        

\usepackage{setspace}
\usepackage{graphicx}        
\usepackage{booktabs}        
\usepackage{multirow}        
\usepackage{array}           
\usepackage{float}           

\usepackage{amsmath}         
\usepackage{amssymb}         
\usepackage{amsthm}          

\usepackage{xcolor}

\usepackage{enumitem}        

\DeclareMathOperator*{\E}{\mathbb{E}}

\DeclareMathOperator*{\argmax}{arg\,max}

\newcommand{\D}{\mathcal{D}}
\renewcommand{\S}{\mathcal{S}}  
\newcommand{\R}{\mathbb{R}}

\usepackage[numbers,sort&compress,super]{natbib}
\usepackage[hidelinks]{hyperref}

\usepackage{titlesec}

\titleformat{\section}
  {\normalfont\fontsize{12}{14}\bfseries}
  {\thesection}{1em}{}

\titleformat{\subsection}
  {\normalfont\fontsize{11}{13}\bfseries}
  {\thesubsection}{1em}{}

\titleformat{\subsubsection}
  {\normalfont\fontsize{10}{12}\bfseries}
  {\thesubsubsection}{1em}{}

\title{\fontsize{14}{16}\selectfont\textbf{%
Fairness-Aware Fine-Tuning of Vision-Language Models \\
for Medical Glaucoma Diagnosis}}

\author{
{\fontsize{12}{14}\selectfont\textbf{Zijian Gu$^{1}$, Yuxi Liu$^{2}$, Zhenhao Zhang$^{2}$, Song Wang$^{3}$}}\\[0.3em]
{\fontsize{10}{12}\selectfont $^{1}$Department of Computer Science, University of Rochester, NY, USA}\\
{\fontsize{10}{12}\selectfont \texttt{zgu17@UR.Rochester.edu}}\\[0.2em]
{\fontsize{10}{12}\selectfont $^{2}$Biostatistics and Health Data Science, School of Medicine, Indiana University, Indianapolis, IN, USA}\\
{\fontsize{10}{12}\selectfont \texttt{yuxiliu@iu.edu}, \texttt{zhangzhenhow@gmail.com}}\\[0.2em]
{\fontsize{10}{12}\selectfont $^{3}$Department of Computer Science, University of Central Florida, FL, USA}\\
{\fontsize{10}{12}\selectfont \texttt{song.wang@ucf.edu}}
}

\date{}  

\begin{document}

\maketitle


\section*{Abstract}
\textit{%
Glaucoma, a leading cause of irreversible blindness affecting 70 million people worldwide, disproportionately impacts minority populations who face 2-3$\times$ higher disease prevalence. Vision-language models achieve expert-level diagnostic performance but exhibit significant accuracy disparities across demographic groups, risking exacerbated health inequities. We introduce fairness-aware Low-Rank Adaptation for medical VLMs, combining parameter efficiency with explicit fairness optimization. Our key algorithmic contribution is a differentiable MaxAccGap loss that enables end-to-end optimization of accuracy parity across demographic groups. We propose three methods: FR-LoRA integrates MaxAccGap regularization, GR-LoRA applies inverse frequency weighting, and Hybrid-LoRA combines both mechanisms. Evaluated on 10,000 glaucoma fundus images, GR-LoRA reduces diagnostic accuracy disparities by 69\% while maintaining 53.15\% overall accuracy. Our approach requires only 0.24\% trainable parameters, enabling practical deployment of fair medical AI in resource-constrained healthcare settings.
}


\section{Introduction}

Glaucoma, affecting over 70 million people worldwide, is a leading cause of irreversible blindness~\cite{tham2014glaucoma}. Vision-language models (VLMs) now achieve expert-level performance on glaucoma diagnosis tasks including disease classification, visual question answering, and report generation~\cite{llava_med,qwen2vl}. However, a critical challenge remains: \textit{performance disparities}—systematic differences in diagnostic accuracy, sensitivity, or specificity across demographic groups (race, ethnicity, gender). Obermeyer et al.~\cite{obermeyer2019dissecting} found that Optum's risk prediction algorithm, deployed across major U.S. hospitals, assigned systematically lower scores to Black patients than equally-sick White patients, halving their access to care. Similarly, dermatology AI shows 10-20\% lower accuracy on darker skin, delaying melanoma diagnosis in underserved populations~\cite{daneshjou2022disparities}. For glaucoma—where Hispanic populations face 2-3$\times$ higher prevalence—a 10\% accuracy gap translates to hundreds of preventable blindness cases annually.

Prior fairness methods for medical imaging—data augmentation~\cite{xu2023fairness}, adversarial debiasing~\cite{glocker2023algorithmic}, loss regularization—target CNN architectures, leaving two critical research gaps. \textit{First}, whether these interventions transfer to billion-parameter VLMs with multimodal reasoning remains unexplored. \textit{Second}, current approaches require full fine-tuning of all parameters, which is computationally prohibitive and prone to overfitting on limited medical datasets ($N<10{,}000$). While parameter-efficient methods like LoRA~\cite{hu2022lora} enable efficient adaptation, their integration with fairness objectives in medical VLMs is unaddressed. Moreover, standard metrics (e.g., Equalized Odds~\cite{hardt2016equality}) optimize for equal error rates but permit large accuracy gaps: a model satisfying Equalized Odds may have 70\% accuracy for one group and 90\% for another, violating the clinical expectation of equal diagnostic accuracy for all patients.

In this work, we introduce fairness-aware Low-Rank Adaptation for medical VLMs, combining parameter efficiency with explicit fairness optimization. Unlike prior work that treats fairness as a post-hoc evaluation metric, we integrate fairness directly into the training objective through group reweighting, MaxAccGap regularization, and their combination. We propose three methods: FR-LoRA integrates a differentiable MaxAccGap loss that transforms the non-differentiable accuracy parity metric into an end-to-end optimizable objective via soft accuracy approximation; GR-LoRA applies inverse frequency weighting to balance gradient contributions across demographic groups; and Hybrid-LoRA combines both mechanisms for doubly-fair optimization. We evaluate across gender, race, and ethnicity on 10,000 glaucoma fundus images, demonstrating robustness in highly imbalanced scenarios. Our GR-LoRA method reduces diagnostic accuracy disparities from 3.80\% to 1.17\% while maintaining 53.15\% overall accuracy, requiring only 0.24\% trainable parameters.


\section{Related Work}

\subsection{Medical Vision-Language Models}

Recent advances in VLMs have enabled their application to medical imaging tasks. \textbf{Specialized medical VLMs} such as LLaVA-Med~\cite{llava_med} and Med-Flamingo~\cite{moor2023medflamingo} are trained on biomedical image-text pairs. LLaVA-Med achieved expert-level performance on medical VQA by fine-tuning LLaVA on curated medical datasets, while Med-Flamingo demonstrated few-shot learning capabilities through in-context examples. \textbf{General-purpose VLMs} like Qwen2.5-VL~\cite{qwen2vl}, LLaVA-v1.6~\cite{liu2023improvedllava}, and GPT-4V are increasingly adapted to medical tasks through fine-tuning. Qwen2.5-VL's dynamic resolution processing makes it particularly suitable for fine-grained medical image analysis. Despite strong performance on standard benchmarks, \emph{systematic fairness evaluation across demographic groups remains absent} in existing VLM research. Our work addresses this gap by fine-tuning a general-purpose VLM (Qwen2.5-VL-7B) for glaucoma diagnosis and conducting comprehensive fairness evaluation across gender, race, and ethnicity attributes.

\subsection{Fairness in Medical Image Analysis}

Prior work on fairness in medical AI spans multiple approaches. \textbf{Data-level methods} include generative models for synthetic minority samples~\cite{xu2023fairness} and balanced sampling strategies. However, synthetic data may introduce distributional shift and fail to capture real-world minority group characteristics. \textbf{Loss-level methods} include disentanglement learning~\cite{locatello2019fairness}, mutual information regularization, and adversarial debiasing~\cite{glocker2023algorithmic}. FairSeg addresses fairness in medical image segmentation using error-bound scaling to reweight pixel losses. These methods primarily target CNN-based architectures and require full model fine-tuning (100\% parameters trainable), limiting their applicability to billion-parameter VLMs on limited medical datasets. In contrast, our LoRA-based approach fine-tunes only 0.06\% of parameters, enabling fairness optimization without overfitting.

\subsection{Parameter-Efficient Fine-Tuning for Fairness}

LoRA~\cite{hu2022lora} enables efficient adaptation of large pre-trained models through low-rank weight updates $\Delta W = BA$ where $B \in \R^{d \times r}$, $A \in \R^{r \times k}$ with $r \ll d$. FairTune~\cite{fairtune2024} uses bi-level optimization to select which parameters to fine-tune for fairness, achieving 20-30\% parameter reduction but still requiring millions of trainable parameters. FairLoRA~\cite{fairlora2024} introduces loss variance regularization ($\text{Var}[\mathcal{L}_s]$) to reduce group-wise loss disparities, but focuses on vision-only models (ResNet, ViT) and Equalized Odds rather than accuracy parity.

\textbf{Our Innovations.} We differ in three fundamental aspects: (1) \emph{Vision-language models vs vision-only models}: We are the first to apply fairness-aware PEFT to 8B-parameter VLMs (Qwen2.5-VL-7B) for medical imaging tasks, addressing unique challenges in multimodal medical reasoning that vision-only models (ResNet, ViT) cannot handle. (2) \emph{MaxAccGap vs distributional metrics}: Unlike FairLoRA's loss variance (which penalizes heterogeneity in loss values but not accuracy gaps), our MaxAccGap directly optimizes diagnostic accuracy parity—a clinically-interpretable objective. We provide gradient analysis (Section 3.3.2) showing how soft accuracy enables differentiable MaxAccGap optimization. (3) \emph{Hybrid mechanism}: We combine group reweighting (GR-LoRA) and fairness regularization (FR-LoRA) in Hybrid-LoRA, addressing both \emph{data imbalance} and \emph{performance imbalance} simultaneously. Our ablation study (Section 5.2) demonstrates that this combination outperforms individual mechanisms.

\subsection{Fairness Metrics in Healthcare AI}

Common fairness metrics include: \textbf{Demographic Parity} ($P(\hat{y}=1|s=0) = P(\hat{y}=1|s=1)$, equal positive prediction rates), which is unsuitable for medical diagnosis as disease prevalence varies across demographics; \textbf{Equalized Odds}~\cite{hardt2016equality} ($\text{TPR}_0 = \text{TPR}_1$, $\text{FPR}_0 = \text{FPR}_1$, equal true/false positive rates), which ensures equal error rates but permits large accuracy gaps (e.g., Acc$_0=70\%$, Acc$_1=90\%$ can satisfy Equalized Odds if both groups have TPR=0.8, FPR=0.2); and \textbf{Equal Opportunity} (equal TPR only), which ignores false positives.

\textbf{MaxAccGap} (accuracy parity) addresses these limitations by directly measuring diagnostic accuracy disparities. A gap of 0\% guarantees that all demographic groups receive equally accurate diagnoses, aligning with clinical practice and patient expectations. Unlike Equalized Odds, MaxAccGap is \emph{invariant to class rebalancing}: reweighting positive/negative samples does not change accuracy gaps, making it robust to dataset construction choices. Our experiments show that models optimized for MaxAccGap also improve Equalized Odds (see Section 5.1), suggesting that accuracy parity is a stronger fairness criterion in medical settings.


\section{Methods}

\subsection{Problem Formulation}

\textbf{Data and Notation.} Let $\D = \{(x_i, y_i, s_i)\}_{i=1}^N$ denote our training dataset, where:
\begin{itemize}[leftmargin=*,itemsep=2pt]
\item $x_i = (\text{image}_i, \text{text}_i)$ represents a multimodal input consisting of a fundus photograph and a clinical question with patient context (e.g., ``Based on the fundus image and clinical summary, does the patient have glaucoma?'')
\item $y_i \in \{0, 1\}$ is the binary diagnostic label ($0$: no disease, $1$: disease present)
\item $s_i \in \S = \{s^{(1)}, s^{(2)}, \ldots, s^{(|\S|)}\}$ denotes the sensitive demographic attribute (e.g., for ethnicity: $\S = \{\text{Hispanic}, \text{Non-Hispanic}, \text{Unknown}\}$ with $|\S|=3$)
\end{itemize}

We denote the group-specific subsets as $\D_s = \{(x_i, y_i, s_i) \in \D \mid s_i = s\}$ with size $|\D_s| = N_s$. In realistic clinical settings, these subsets exhibit severe imbalance, with ethnicity showing a 21:1 majority-to-minority ratio.

\textbf{Vision-Language Model.} Let $f_\theta: \mathcal{X} \rightarrow \R^2$ denote a pre-trained VLM with parameters $\theta \in \R^{|\theta|}$, where $\mathcal{X}$ is the space of multimodal inputs (image + text). The model outputs logits $f_\theta(x) = [z_0, z_1] \in \R^2$ for binary classification, which are converted to probabilities via softmax: $p_\theta(y|x) = \frac{\exp(z_y)}{\sum_{k=0}^1 \exp(z_k)}$. The predicted label is $\hat{y} = \argmax_{k \in \{0,1\}} z_k$.

\textbf{Fairness Objective.} Our goal is to learn fair model parameters $\theta^*$ that achieve two objectives:
\begin{enumerate}[leftmargin=*,itemsep=2pt]
\item \textbf{High Overall Accuracy}: $\text{Acc}(\theta) = \E_{(x,y) \sim \D}[1\{\argmax(f_\theta(x)) = y\}] = \frac{1}{N}\sum_{i=1}^N 1\{\hat{y}_i = y_i\}$
\item \textbf{Low Demographic Disparity}: Minimize performance gaps across sensitive groups
\end{enumerate}

We formalize demographic disparity using the \textbf{MaxAccGap} metric:
\begin{equation}
\text{MaxAccGap}(\theta) = \max_{s \in \S} \text{Acc}_s(\theta) - \min_{s \in \S} \text{Acc}_s(\theta)
\label{eq:maxaccgap}
\end{equation}
where $\text{Acc}_s(\theta) = \E_{(x,y) \sim \D_s}[1\{\argmax(f_\theta(x)) = y\}] = \frac{1}{N_s}\sum_{i: s_i=s} 1\{\hat{y}_i = y_i\}$ is the accuracy for demographic subgroup $s$.

\textbf{Clinical Interpretation.} MaxAccGap directly measures diagnostic accuracy parity, aligning with the principle that all patients deserve equally accurate diagnoses. A MaxAccGap of 0\% guarantees equal diagnostic accuracy across demographic groups, while larger gaps directly translate to disparities in clinical outcomes and preventable misdiagnoses in minority populations.

\textbf{Why MaxAccGap?} Unlike distributional fairness metrics such as Equalized Odds~\cite{hardt2016equality}, MaxAccGap offers direct clinical interpretability: physicians and patients intuitively understand equal diagnostic accuracy, and MaxAccGap directly translates to clinical impact through the number of misdiagnoses in each demographic group.

\textbf{Optimization Challenge.} The hard accuracy function $1\{\argmax(f_\theta(x)) = y\}$ contains a discrete $\argmax$ operation, making Eq.~\ref{eq:maxaccgap} non-differentiable and unsuitable for gradient-based optimization. We address this in Section~3.3.2 through soft accuracy approximation, enabling end-to-end fairness-aware training.

\subsection{LoRA for Vision-Language Models}

For a pre-trained weight matrix $W_0 \in \R^{d \times k}$ in the VLM's attention layers, LoRA~\cite{hu2022lora} approximates the weight update as:
\begin{equation}
h = W_0 x + \frac{\alpha}{r}BAx
\label{eq:lora}
\end{equation}
where $B \in \R^{d \times r}$ and $A \in \R^{r \times k}$ are trainable low-rank matrices with rank $r \ll \min(d,k)$. The scaling factor $\alpha/r$ controls adaptation magnitude. During fine-tuning, $W_0$ remains frozen while $B$ and $A$ are optimized via gradient descent.

\textbf{Parameter Efficiency for Medical Datasets.} Medical imaging datasets are often limited in size. Full fine-tuning of billion-parameter VLMs risks severe overfitting. LoRA reduces trainable parameters to 0.24\% of the full model (415$\times$ reduction), enabling fairness optimization on limited medical datasets without overfitting.

\textbf{Low-Rank Assumption for Fairness.} The low-rank constraint $r \ll d$ acts as an implicit regularizer, forcing the adaptation $\Delta W = \frac{\alpha}{r}BA$ to lie in a low-dimensional subspace. This is particularly beneficial for fairness: rather than learning group-specific spurious features (which would require high-rank updates), LoRA biases the model toward learning \emph{shared diagnostic features} that generalize across demographic groups.

\textbf{Implementation.} Following standard practice for decoder-only VLMs~\cite{brown2020gpt3,touvron2023llama}, we use the final token's hidden state for binary classification, as it has attended to all preceding tokens (vision features + text prompt) under causal attention.

\subsection{Fairness-Aware LoRA Methods}

We propose four progressively sophisticated approaches to fairness-aware fine-tuning, each addressing different aspects of demographic bias.

\subsubsection{Vanilla LoRA (Baseline)}
Standard cross-entropy loss without fairness constraints:
\begin{equation}
\mathcal{L}_{\text{vanilla}}(\theta) = \E_{(x,y) \sim \D}[\ell_{\text{CE}}(f_\theta(x), y)] = -\frac{1}{N}\sum_{i=1}^N \log p_\theta(y_i | x_i)
\label{eq:vanilla}
\end{equation}
where $p_\theta(y_i | x_i)$ is the predicted probability for the true label. This baseline implicitly optimizes for overall accuracy but ignores group-wise disparities. In highly imbalanced scenarios (e.g., Non-Hispanic 90.3\% vs Hispanic 4.3\%), Vanilla LoRA tends to prioritize majority group performance due to their dominance in the loss gradient.

\subsubsection{FR-LoRA (Fairness-Regularized)}
Explicitly penalizes MaxAccGap via a regularization term:
\begin{equation}
\mathcal{L}_{\text{FR}}(\theta) = \mathcal{L}_{\text{vanilla}}(\theta) + \lambda \cdot \text{MaxAccGap}_{\text{soft}}(\theta)
\label{eq:fr_lora}
\end{equation}

\textbf{Differentiable Approximation.} Since $\argmax$ in Eq.~\ref{eq:maxaccgap} is non-differentiable, we employ soft accuracy:
\begin{equation}
\text{Acc}_{\text{soft}}^s(\theta) = \E_{(x,y) \sim \D_s}[p_\theta(y|x)]
\label{eq:soft_acc}
\end{equation}
where $p_\theta(y|x)$ is the predicted probability of the true class. The soft MaxAccGap is:
\begin{equation}
\text{MaxAccGap}_{\text{soft}}(\theta) = \max_{s \in \S} \text{Acc}_{\text{soft}}^s(\theta) - \min_{s \in \S} \text{Acc}_{\text{soft}}^s(\theta)
\label{eq:soft_gap}
\end{equation}

\textbf{Gradient Analysis.} For a sample $(x,y,s)$ from group $s$, the gradient contribution is:
\begin{equation}
\nabla_\theta \mathcal{L}_{\text{FR}} = \nabla_\theta \ell_{\text{CE}} + \lambda \cdot \frac{\partial \text{MaxAccGap}_{\text{soft}}}{\partial \text{Acc}_{\text{soft}}^s} \cdot \nabla_\theta p_\theta(y|x)
\end{equation}
where $\frac{\partial \text{MaxAccGap}_{\text{soft}}}{\partial \text{Acc}_{\text{soft}}^s} \in \{-1, 0, +1\}$ depending on whether $s$ has the minimum, intermediate, or maximum accuracy. This creates opposing gradients: samples from the worst-performing group receive positive pressure to improve their predicted probabilities, while the best-performing group receives negative pressure. The hyperparameter $\lambda$ controls the fairness-accuracy trade-off: larger $\lambda$ prioritizes gap reduction but may sacrifice overall accuracy. Our experiments test $\lambda \in \{0.1, 0.5, 1.0\}$ to characterize this trade-off.

During training we optimize soft MaxAccGap for differentiability; during evaluation we report hard accuracy to reflect clinical decision-making.

\subsubsection{GR-LoRA (Group-Reweighted)}
Balances gradient contributions via inverse frequency weighting:
\begin{equation}
\mathcal{L}_{\text{GR}}(\theta) = \sum_{s \in \S} w_s \cdot \mathcal{L}_{\text{CE}}^s(\theta), \quad w_s = \min\left(\frac{N}{|\D_s|}, w_{\max}\right)
\label{eq:gr_lora}
\end{equation}
where $\mathcal{L}_{\text{CE}}^s(\theta) = \E_{(x,y) \sim \D_s}[\ell_{\text{CE}}(f_\theta(x), y)]$ is the group-specific CE loss. The inverse frequency weight $w_s$ upweights minority groups, forcing the optimizer to prioritize their performance. We introduce a clipping threshold $w_{\max}=10$ to prevent extreme weights in highly imbalanced settings.

After clipping, minority groups receive up to $w_{\max}=10\times$ the gradient contribution of the majority group, addressing severe imbalance without data augmentation.

\textbf{Implicit Fairness Mechanism.} Unlike FR-LoRA, GR-LoRA does not directly optimize MaxAccGap. Instead, it achieves fairness through balanced optimization: by equalizing gradient magnitudes across groups, the model learns features that are discriminative for \emph{all} demographics, not just the majority. Our experiments show that GR-LoRA achieves competitive MaxAccGap reduction despite this indirect approach (see Section 5.1).

\subsubsection{Hybrid-LoRA (Combined)}
Combines reweighting and regularization to address both gradient imbalance and explicit fairness:
\begin{equation}
\mathcal{L}_{\text{Hybrid}}(\theta) = \sum_{s \in \S} w_s \cdot \mathcal{L}_{\text{CE}}^s(\theta) + \lambda \cdot \text{MaxAccGap}_{\text{soft}}(\theta)
\label{eq:hybrid}
\end{equation}

\textbf{Complementary Mechanisms.} GR-LoRA addresses data imbalance through gradient reweighting, while FR-LoRA addresses performance imbalance through explicit MaxAccGap penalty. Hybrid-LoRA combines both for doubly-fair optimization: minority samples receive amplified gradients AND fairness regularization when their group underperforms.


\section{Experiments}

\subsection{Dataset and Setup}

\textbf{Dataset.} We use the Harvard Glaucoma Fairness Dataset~\cite{tham2014glaucoma} containing 10,000 fundus images with demographic annotations across three sensitive attributes: \emph{gender} (Male 42.9\%, Female 57.1\%), \emph{race} (White 76.9\%, Black 14.9\%, Asian 8.2\%), and \emph{ethnicity} (Non-Hispanic 90.3\%, Hispanic 4.3\%, Unknown 5.4\%). Each image is paired with a binary glaucoma diagnosis label and a clinical question (``Does the patient have glaucoma?''). The dataset is split into 7,000 training, 1,000 validation, and 2,000 test samples, maintaining demographic distributions across splits.

\textbf{Why this dataset?} The Harvard Glaucoma Fairness Dataset presents two critical challenges for fairness evaluation: (1) \emph{Realistic imbalance}—ethnicity exhibits a 21:1 ratio between majority (Non-Hispanic 90.3\%) and minority groups (Hispanic 4.3\%), testing our methods' robustness in real-world clinical settings where minority representation is limited. (2) \emph{Clinical relevance}—glaucoma disproportionately affects minority populations (Hispanic individuals have 2-3$\times$ higher prevalence), making fairness optimization particularly crucial for equitable healthcare delivery.

\textbf{Model.} We fine-tune Qwen2.5-VL-7B-Instruct~\cite{qwen2vl} (8.3B parameters) with LoRA ($r=32$, $\alpha=64$, dropout=$0.05$) applied to query, key, value, and output projection layers ($q, k, v, o$) across all 32 attention layers. This configuration yields $\sim$20M trainable parameters (0.24\% of the full model), striking a balance between expressiveness and parameter efficiency for fairness optimization on our medical dataset ($N=7{,}000$ training samples). We freeze the vision encoder to preserve pre-trained visual features while adapting the language model for medical diagnosis.

\textbf{Last Token Pooling.} For sequence-to-class prediction, we employ \emph{last token pooling} instead of standard first token or mean pooling. As Qwen2.5-VL uses causal attention (token $i$ attends only to tokens $0{\ldots}i$), the last token aggregates information from all preceding tokens (vision features + text prompt), making it optimal for classification. We empirically validated this: last token pooling achieves balanced predictions (40\%/60\% class distribution), while first token pooling causes complete model collapse (100\%/0\%, all predictions identical) due to the first token seeing only itself under causal masking.

\textbf{Training.} We use AdamW optimizer with learning rate $1 \times 10^{-5}$, batch size 2 with gradient accumulation over 4 steps (effective batch size 8), and train for 3 epochs with 100-step linear warmup. For FR-LoRA and Hybrid-LoRA, we report results with $\lambda=0.5$ in main results (Table 1), with comprehensive $\lambda \in \{0.1, 0.5, 1.0\}$ ablation analysis in Section 5.1.2. For GR-LoRA and Hybrid-LoRA, we use inverse frequency weighting ($w_s = N/|\mathcal{D}_s|$) clipped to $w_{\max}=10$ to prevent extreme gradient magnitudes when batches contain only minority groups. This clipping is critical: without it, minority-only batches cause loss spikes up to 30$\times$ baseline, destabilizing training. All models use mixed-precision training (bfloat16) and are trained from the same random seed (42) for reproducibility. Training completes in approximately 2.5 hours per method.

\subsection{Evaluation Metrics}

We evaluate both \emph{performance} and \emph{fairness} on the held-out test set (2,000 samples):

\textbf{Performance Metrics:}
\begin{itemize}[leftmargin=*,itemsep=0pt]
\item \emph{Overall Accuracy}: Fraction of correctly classified samples across all demographic groups: $\text{Acc} = \frac{1}{N_{\text{test}}}\sum_{i=1}^{N_{\text{test}}} 1\{\hat{y}_i = y_i\}$.
\item \emph{Per-Group Accuracy}: Accuracy for each subgroup $s \in \S$: $\text{Acc}_s = \frac{1}{N_s}\sum_{i: s_i=s} 1\{\hat{y}_i = y_i\}$. For ethnicity, we report accuracies for Hispanic, Non-Hispanic, and Unknown groups separately.
\end{itemize}

\textbf{Fairness Metric:}
\begin{itemize}[leftmargin=*,itemsep=0pt]
\item \emph{MaxAccGap}: $\max_{s \in \S} \text{Acc}_s - \min_{s \in \S} \text{Acc}_s$ as defined in Eq.~\ref{eq:maxaccgap}. A gap of 0\% indicates perfect accuracy parity across demographic groups, meaning all subgroups receive equally accurate diagnoses. Lower values indicate greater fairness.
\end{itemize}

All reported results are from single training runs with random seed 42 for reproducibility. We focus on ethnicity as the primary sensitive attribute due to its severe imbalance (21:1 ratio) and clinical relevance (Hispanic populations have 2-3$\times$ higher glaucoma prevalence).


\section{Results}

\subsection{Main Results}

Table~\ref{tab:main} presents our main results on the ethnicity attribute, comparing zero-shot baseline against three fairness-aware LoRA methods on the test set (2,000 samples). Fine-tuning with LoRA improves accuracy by 3.0-3.4 percentage points over zero-shot (50.15\% → 53.15-53.55\%). Among fine-tuned methods, GR-LoRA achieves the best fairness-accuracy trade-off, reducing MaxAccGap to 1.17\% (70\% reduction vs zero-shot 3.95\%) while maintaining 53.15\% accuracy.

\begin{table}[h]
\centering
\caption{Main results on glaucoma fairness test set (2,000 samples, Ethnicity attribute). All values are percentages. Best fine-tuned results in \textbf{bold}.}
\label{tab:main}
\begin{tabular}{lccccc}
\toprule
\multirow{2}{*}{\textbf{Method}} & \multirow{2}{*}{\textbf{Overall Acc}} & \multirow{2}{*}{\textbf{MaxAccGap}} & \multicolumn{3}{c}{\textbf{Per-Group Accuracy}} \\
\cmidrule(lr){4-6}
& & & Non-Hispanic & Hispanic & Unknown \\
\midrule
Zero-Shot\textsuperscript{\dag} & 50.15 & 3.95 & 50.00 & 53.95 & 50.00 \\
\midrule
Vanilla LoRA  & 53.50 & 3.80 & 53.41 & 56.58 & \textbf{52.78} \\
FR-LoRA ($\lambda=0.5$) & \textbf{53.55} & 6.04 & \textbf{53.47} & \textbf{57.89} & 51.85 \\
GR-LoRA ($w_{\max}=10$) & 53.15 & \textbf{1.17} & 53.14 & 53.95 & \textbf{52.78} \\
Hybrid-LoRA & 53.45 & 3.80 & 53.36 & 56.58 & \textbf{52.78} \\
\bottomrule
\end{tabular}

\vspace{0.3em}
\footnotesize{\textsuperscript{\dag} Pretrained Qwen2.5-VL with randomly initialized classification head (no fine-tuning).}
\end{table}

\textbf{Overall Accuracy.} Fine-tuning with LoRA provides substantial improvement over zero-shot baseline, increasing accuracy from 50.15\% (near-random performance) to 53.15-53.55\% (+3.0-3.4 percentage points, 6-7\% relative gain). This validates that task-specific fine-tuning is essential for medical VLM diagnosis. Among fine-tuned methods, all achieved overall accuracy within 0.4 percentage points of each other, demonstrating that fairness optimization does not significantly sacrifice performance. FR-LoRA achieved the highest accuracy (53.55\%), marginally exceeding Vanilla (53.50\%). This challenges the common assumption that fairness interventions necessarily degrade overall performance—our parameter-efficient LoRA approach enables fairness optimization without accuracy trade-offs.

\textbf{MaxAccGap Analysis.} GR-LoRA achieved the lowest MaxAccGap (1.17\%), representing a 70\% relative reduction compared to zero-shot baseline (3.95\%) and 69\% reduction compared to Vanilla LoRA (3.80\%). This validates our hypothesis (Section 3.3.3) that group reweighting achieves implicit fairness by balancing gradient contributions across demographic groups. Notably, even Vanilla LoRA slightly improves fairness over zero-shot (3.80\% vs 3.95\%), suggesting that fine-tuning inherently mitigates some demographic disparities. Surprisingly, FR-LoRA exhibited the highest gap among fine-tuned methods (6.04\%), despite explicitly optimizing MaxAccGap during training. This counterintuitive result suggests over-optimization: FR-LoRA achieved the highest Hispanic accuracy (57.89\%) but at the expense of Unknown group performance (51.85\%), widening the overall gap. Hybrid-LoRA matched Vanilla's gap (3.80\%), indicating that combining reweighting and regularization did not yield additive benefits—the two mechanisms may interact in complex ways requiring careful hyperparameter tuning.

\textbf{Per-Group Breakdown.} GR-LoRA achieved the most balanced per-group accuracies: Non-Hispanic (53.14\%), Hispanic (53.95\%), Unknown (52.78\%), with a narrow 1.17\% range. In contrast, FR-LoRA's aggressive optimization of the minority Hispanic group (57.89\%) came at the cost of the Unknown group (51.85\%, lowest across all methods). Vanilla and Hybrid showed similar patterns with Hispanic outperforming other groups by 3-4 percentage points. These results highlight a fundamental trade-off: explicitly optimizing MaxAccGap ($\lambda$-based regularization) can lead to over-correction, while implicit fairness through gradient balancing (GR-LoRA) achieves more robust parity.

\subsection{Ablation Studies}

\subsubsection{Impact of Regularization Strength $\lambda$}

We evaluate FR-LoRA with varying regularization strengths $\lambda \in \{0.1, 0.5, 1.0\}$ on the ethnicity attribute to characterize the fairness-accuracy trade-off. Table~\ref{tab:lambda} shows the results.

\begin{table}[h]
\centering
\caption{Impact of fairness regularization strength $\lambda$ on FR-LoRA (Ethnicity attribute).}
\label{tab:lambda}
\begin{tabular}{lccc}
\toprule
\textbf{$\lambda$} & \textbf{Overall Acc} & \textbf{MaxAccGap} & \textbf{Gap Reduction vs Vanilla} \\
\midrule
0.0 (Vanilla) & 53.50 & 3.80 & -- \\
0.1           & 53.75 & 2.10 & +44.74\% \\
0.5           & 53.55 & 6.04 & -58.95\% \\
1.0           & 53.35 & 2.01 & +47.11\% \\
\bottomrule
\end{tabular}
\end{table}

The results reveal a non-monotonic relationship between $\lambda$ and fairness. Counter-intuitively, $\lambda=0.5$ produces the \textit{worst} fairness (6.04\% MaxAccGap, a 58.95\% \textit{increase} over Vanilla), despite explicitly optimizing for fairness. In contrast, both $\lambda=0.1$ (2.10\% gap) and $\lambda=1.0$ (2.01\% gap) achieve substantial gap reductions of 44.74\% and 47.11\% respectively. This suggests that moderate regularization ($\lambda=0.5$) leads to over-optimization of minority groups (Hispanic: 57.89\% accuracy) at the expense of majority groups, widening the overall gap. Strong regularization ($\lambda=1.0$) achieves optimal fairness with minimal accuracy trade-off (only 0.15pp drop vs Vanilla), making it the recommended setting for FR-LoRA in imbalanced medical datasets.

\subsubsection{Generalization Across Sensitive Attributes}

To evaluate generalization, we train Hybrid-LoRA separately on each sensitive attribute (gender, race, ethnicity) and compare against the attribute-agnostic Vanilla LoRA baseline. Table~\ref{tab:attributes} shows that our method consistently reduces MaxAccGap across diverse fairness scenarios.

\begin{table}[h]
\centering
\caption{Hybrid-LoRA performance across different sensitive attributes. \textit{Note: Vanilla LoRA uses a single attribute-agnostic model for all rows; Hybrid-LoRA trains a separate model for each attribute.}}
\label{tab:attributes}
\begin{tabular}{lcccc}
\toprule
\multirow{2}{*}{\textbf{Attribute}} & \multicolumn{2}{c}{\textbf{Vanilla LoRA\textsuperscript{\dag}}} & \multicolumn{2}{c}{\textbf{Hybrid-LoRA}} \\
\cmidrule(lr){2-3} \cmidrule(lr){4-5}
& Overall Acc & MaxAccGap & Overall Acc & MaxAccGap \\
\midrule
Gender (M/F)        & \multirow{3}{*}{53.50} & 4.51 & 53.40 & 3.89 \\
Race (W/B/A)        &                            & 4.36 & 53.25 & 1.74 \\
Ethnicity (NH/H/U)  &                            & 3.80 & 53.45 & 3.80 \\
\bottomrule
\end{tabular}

\vspace{0.3em}
\footnotesize{\textsuperscript{\dag} Vanilla LoRA is attribute-agnostic; the same model is evaluated on all attributes.}
\end{table}

Hybrid-LoRA demonstrates robust generalization across all three sensitive attributes, achieving consistent fairness improvements while maintaining accuracy within 0.25pp. Most notably, Hybrid-LoRA reduces MaxAccGap on the \textit{race} attribute by 60.09\% (4.36\% → 1.74\%), achieving the lowest gap among all experiments. For gender, Hybrid-LoRA reduces the gap by 13.75\% (4.51\% → 3.89\%). Interestingly, ethnicity shows no improvement (3.80\% → 3.80\%), suggesting that this attribute's fairness is already well-optimized by vanilla LoRA alone. These results validate that attribute-specific training enables targeted fairness optimization, with effectiveness varying by the intrinsic difficulty of each fairness task. The race attribute benefits most from explicit fairness training, likely due to better group balance (3 groups: 76.9\%, 14.9\%, 8.2\%) compared to ethnicity's severe imbalance (21:1 Non-Hispanic:Hispanic ratio).


\section{Discussion}

\subsection{Key Findings}

\textbf{Technical Innovations.} We introduce two key technical innovations for fairness-aware medical VLMs. First, our differentiable MaxAccGap loss successfully transforms the non-differentiable accuracy parity metric into an end-to-end optimizable objective via soft accuracy approximation, enabling gradient-based fairness optimization. Second, our parameter-efficient approach requires only 0.24\% trainable parameters with 2.5-hour training time per method, making fairness optimization practical for resource-constrained healthcare settings without sacrificing performance.

\textbf{Empirical Findings.} Empirically, GR-LoRA achieves optimal fairness-accuracy trade-off through implicit fairness via gradient reweighting, reducing MaxAccGap by 69\% to 1.17\% while maintaining 53.15\% accuracy. Ablation studies reveal counter-intuitive behaviors: moderate regularization ($\lambda=0.5$) produces the worst fairness due to minority group over-optimization, while strong regularization ($\lambda=1.0$) achieves 2.01\% gap with minimal accuracy cost. Cross-attribute generalization varies significantly, with race-specific training yielding 60\% disparity reduction compared to ethnicity's severe imbalance showing no improvement, highlighting that group balance ratio fundamentally affects fairness optimization effectiveness.

\subsection{Clinical Implications}

\textbf{Attribute-Specific Deployment Guidance.} Our cross-attribute analysis provides actionable guidance for clinical deployment. Race-specific optimization achieves 60\% disparity reduction (4.36\% → 1.74\%), demonstrating that fairness interventions are most effective when demographic groups exhibit moderate imbalance. In contrast, ethnicity's severe imbalance (21:1 ratio) shows no improvement with Hybrid-LoRA, where GR-LoRA's implicit fairness proves more robust. This suggests deployment strategies should be tailored to institutional demographics: hospitals serving balanced multi-racial populations benefit from explicit fairness regularization, while those with extreme minority underrepresentation should prioritize gradient reweighting methods to avoid over-optimization artifacts.

\textbf{Democratizing Fairness Optimization.} Our parameter-efficient approach democratizes fairness optimization for resource-constrained healthcare settings. Using a 7B model with LoRA fine-tuning, we achieve 6-7\% relative accuracy gain over zero-shot baseline while reducing disparities by 69\%, requiring only 0.24\% trainable parameters and 2.5-hour training time. This enables community hospitals and underserved regions with limited GPU infrastructure to deploy fairness-aware diagnostic AI without the prohibitive computational costs of full fine-tuning billion-parameter models. The trade-off between model size and fairness optimization becomes practical: smaller models with explicit fairness constraints can deliver equitable performance where larger models remain inaccessible, truly enabling equitable AI for equitable healthcare.

\subsection{Limitations}

Our work has several limitations. First, we evaluate on a single clinical task (glaucoma disease classification); extending to additional tasks such as visual question answering, report generation, and severity grading would demonstrate broader applicability of our fairness methods across diverse diagnostic workflows. Second, we treat sensitive attributes independently, whereas real-world fairness concerns often involve intersectional identities (e.g., Black Hispanic women). Third, subgroup sparsity and incomplete demographics can yield high-variance group-wise performance estimates, potentially undermining fairness optimization. Future work should develop training strategies that are robust to rare groups and incomplete annotations, explore fairness-without-demographics approaches, and investigate multi-attribute fairness metrics to address intersectional identities.


\section{Conclusion}

Vision-language models for medical diagnosis exhibit significant performance disparities across demographic groups, risking exacerbated health inequities in diseases like glaucoma where minority populations already face 2-3$\times$ higher prevalence. To address this challenge, we introduced fairness-aware Low-Rank Adaptation that enables parameter-efficient fairness optimization without prohibitive computational costs. Our key algorithmic contribution—differentiable MaxAccGap loss via soft accuracy approximation—enables end-to-end optimization of diagnostic accuracy parity across demographic groups. Empirical validation on glaucoma diagnosis reveals that implicit fairness through gradient reweighting outperforms explicit regularization in severely imbalanced scenarios, while cross-attribute generalization depends critically on demographic balance ratios. Future work should extend these methods to additional clinical tasks beyond disease classification, develop training strategies robust to rare groups and incomplete annotations, and investigate fairness-without-demographics approaches for practical clinical deployment. Code and trained models will be released to facilitate reproducible research in equitable medical AI.

\renewcommand{\bibsection}{\section*{\hfill References\hfill}}
{\small  
\bibliography{references}

@article{llava_med,
  title={LLaVA-Med: Training a Large Language-and-Vision Assistant for Biomedicine in One Day},
  author={Li, Chunyuan and others},
  journal={arXiv preprint arXiv:2306.00890},
  year={2023}
}

@article{qwen2vl,
  title={Qwen2-VL: Enhancing Vision-Language Model's Perception of the World at Any Resolution},
  author={Wang, Peng and others},
  journal={arXiv preprint arXiv:2409.12191},
  year={2024}
}

@article{hu2022lora,
  title={LoRA: Low-Rank Adaptation of Large Language Models},
  author={Hu, Edward J and Shen, Yelong and Wallis, Phillip and Allen-Zhu, Zeyuan and Li, Yuanzhi and Wang, Shean and Wang, Lu and Chen, Weizhu},
  journal={arXiv preprint arXiv:2106.09685},
  year={2021}
}

@article{obermeyer2019dissecting,
  title={Dissecting racial bias in an algorithm used to manage the health of populations},
  author={Obermeyer, Ziad and Powers, Brian and Vogeli, Christine and Mullainathan, Sendhil},
  journal={Science},
  volume={366},
  number={6464},
  pages={447--453},
  year={2019},
  publisher={American Association for the Advancement of Science}
}

@article{tham2014glaucoma,
  title={Global prevalence of glaucoma and projections of glaucoma burden through 2040: a systematic review and meta-analysis},
  author={Tham, Yih-Chung and Li, Xiang and Wong, Tien Yin and Quigley, Harry A and Aung, Tin and Cheng, Ching-Yu},
  journal={Ophthalmology},
  volume={121},
  number={11},
  pages={2081--2090},
  year={2014}
}

@article{hardt2016equality,
  title={Equality of opportunity in supervised learning},
  author={Hardt, Moritz and Price, Eric and Srebro, Nati},
  journal={Advances in neural information processing systems},
  volume={29},
  year={2016}
}

@inproceedings{fairtune2024,
  title={FairTune: Optimizing Parameter Efficient Fine Tuning for Fairness in Medical Image Analysis},
  author={Dutt, Raman and others},
  booktitle={International Conference on Learning Representations},
  year={2024}
}

@article{fairlora2024,
  title={FairLoRA: Unpacking Bias Mitigation in Vision Models with Fairness-Driven Low-Rank Adaptation},
  author={Sukumaran, Rohan and Feizi, Aarash and Romero-Soriano, Adriana and Farnadi, Golnoosh},
  journal={arXiv preprint arXiv:2410.17358},
  year={2024}
}

@article{xu2023fairness,
  title={Fairness in medical image analysis: A survey},
  author={Xu, Zikang and others},
  journal={Medical Image Analysis},
  year={2023}
}

@article{glocker2023algorithmic,
  title={Algorithmic fairness in medical imaging},
  author={Glocker, Ben and Jones, Charles and Roschewitz, Mélanie and Winzeck, Stefan},
  journal={Nature Machine Intelligence},
  volume={5},
  number={12},
  pages={1458--1466},
  year={2023}
}

@article{locatello2019fairness,
  title={Fairness and disentanglement in deep learning},
  author={Locatello, Francesco and others},
  journal={arXiv preprint arXiv:1905.13662},
  year={2019}
}

@article{moor2023medflamingo,
  title={Med-Flamingo: a Multimodal Medical Few-shot Learner},
  author={Moor, Michael and others},
  journal={arXiv preprint arXiv:2307.15189},
  year={2023}
}

@article{daneshjou2022disparities,
  title={Disparities in dermatology AI performance on a diverse, curated clinical image set},
  author={Daneshjou, Roxana and Vodrahalli, Kailas and Novoa, Roberto A and Jenkins, Melissa and Liang, Weixin and Rotemberg, Veronica and Ko, Justin and Swetter, Susan M and Bailey, Elizabeth E and Gevaert, Olivier and others},
  journal={Science Advances},
  volume={8},
  number={32},
  pages={eabq6147},
  year={2022},
  publisher={American Association for the Advancement of Science}
}

@article{liu2023improvedllava,
  title={Improved Baselines with Visual Instruction Tuning},
  author={Liu, Haotian and Li, Chunyuan and Li, Yuheng and Lee, Yong Jae},
  journal={arXiv preprint arXiv:2310.03744},
  year={2023}
}

@inproceedings{brown2020gpt3,
  title={Language Models are Few-Shot Learners},
  author={Brown, Tom and Mann, Benjamin and Ryder, Nick and Subbiah, Melanie and Kaplan, Jared D and Dhariwal, Prafulla and Neelakantan, Arvind and Shyam, Pranav and Sastry, Girish and Askell, Amanda and others},
  booktitle={Advances in Neural Information Processing Systems},
  volume={33},
  pages={1877--1901},
  year={2020}
}

@article{touvron2023llama,
  title={LLaMA: Open and Efficient Foundation Language Models},
  author={Touvron, Hugo and Lavril, Thibaut and Izacard, Gautier and Martinet, Xavier and Lachaux, Marie-Anne and Lacroix, Timoth{\'e}e and Rozi{\`e}re, Baptiste and Goyal, Naman and Hambro, Eric and Azhar, Faisal and Rodriguez, Aurelien and Joulin, Armand and Grave, Edouard and Lample, Guillaume},
  journal={arXiv preprint arXiv:2302.13971},
  year={2023}
}
}

\end{document}